\documentclass{article}
\usepackage{ijcai25}

\usepackage{times}
\usepackage{soul}
\usepackage{url}
\usepackage[hidelinks]{hyperref}
\usepackage[utf8]{inputenc}
\usepackage[small]{caption}
\usepackage{graphicx}
\usepackage{amsmath}
\usepackage{amsthm}
\usepackage{booktabs}
\usepackage{algorithm}
\usepackage{algorithmic}
\usepackage[switch]{lineno}
\usepackage{comment}
\usepackage{algorithm}
\usepackage{algorithmic}
\usepackage{multirow}
\usepackage{multicol}
\usepackage{amsmath}

\DeclareMathOperator*{\argmin}{argmin}
\usepackage[table]{xcolor}
\newcommand{\ignore}[1]{}

\usepackage{tabularx}
\usepackage{xcolor}

\definecolor{headergray}{RGB}{240,240,240}
\title{Distributed Multi-Agent Coordination Using Multi-Modal Foundation Models} 

\author{
    Author Name
    \affiliations
    Affiliation
    \emails
    email@example.com
}

\author{
Saaduddin Mahmud
\and
Dorian Benhamou Goldfajn\and
Shlomo Zilberstein
\affiliations
University of Massachusetts Amherst\\
\emails
\{smahmud, dbenhamougol, shlomo\}@umass.edu
}

\begin{document}

\maketitle

\begin{abstract}
Distributed Constraint Optimization Problems (DCOPs) offer a powerful framework for multi-agent coordination but often rely on labor-intensive, manual problem construction. To address this, we introduce {VL-DCOPs}, a framework that takes advantage of large multimodal foundation models (LFMs) to automatically generate constraints from both visual and linguistic instructions. We then introduce a spectrum of agent archetypes for solving VL-DCOPs: from a neuro-symbolic agent that delegates some of the algorithmic decisions to an LFM, to a fully neural agent that depends entirely on an LFM for coordination. We evaluate these agent archetypes using state-of-the-art LLMs (large language models) and VLMs (vision language models) on three novel VL-DCOP tasks and compare their respective advantages and drawbacks. Lastly, we discuss how this work extends to broader frontier challenges in the DCOP literature.
\end{abstract}

\section{Introduction}
Distributed Constraint Optimization Problems (DCOPs) \cite{yokoo1998distributed} are a widely studied framework that has previously been shown to be effective in various applications, including smart homes, robotic and drone coordination, surveillance, economic dispatch, network coordination, and disaster response~\cite{Maheswaran2004TakingDT,farinelli2014agent,zivan2015distributed,fioretto2017distributed,mahmud2020population}. However, in many of these cases, the problems are static and manually modeled, reducing their applicability in changing scenarios. Although some previous works have considered a dynamic formulation of DCOPs~\cite{ddcop}, these approaches still require manual defining of a dynamics model, which limits their adaptability.

Another approach to mitigating this issue is to use preference elicitation to handle partially defined constraints, allowing agents to query humans to reduce uncertainty over time~\cite{Tabakhi2017PreferenceEF,PE2,PEBC,tsouros2024learning,exante}. Even then, initial constraints must be established beforehand, queries often use language templates, and answers to queries take the form of numeric constraint cost values, which may not be a natural or convenient way for humans to communicate their preferences \cite{tversky1982judgment}.

\ignore{
\begin{figure}[t!]
    \centering
    \includegraphics[width=3.3in]{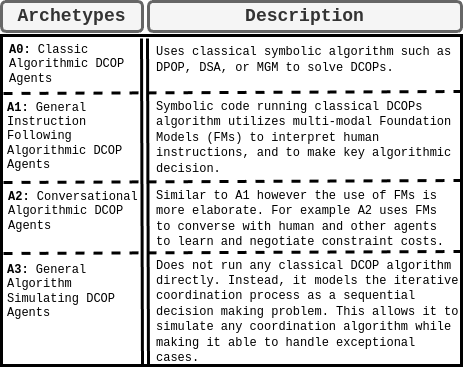}
    \caption{Spectrum of VL-DCOP Agents.}
    \label{fig:SVD}
\end{figure}
}

\begin{table}[t!]
\renewcommand{\arraystretch}{1.2}
\setlength{\tabcolsep}{3pt}
\resizebox{3.375in}{!}{%
\begin{tabular}{|p{1.2in}|p{2.75in}|}
\hline
\cellcolor{lightgray} \textbf{\Large ~~~Archetypes} & \cellcolor{lightgray} \textbf{\Large ~~~Description} \\ \hline \hline
\textbf{A0:} \raggedright Classic Algorithmic DCOP Agents & Agents use classical symbolic algorithms such as DPOP, DSA, or MGM to solve DCOPs. \\ \hline
\textbf{A1:} \raggedright Instruction Following DCOP Agents & Symbolic code executing classical DCOP algorithms leverages LFMs to interpret human instructions and facilitate key algorithmic decisions. \\ \hline
\textbf{A2:} \raggedright Conversational DCOP Agents & Similar to A1, but with a more advanced use of LFMs. A2 employs LFMs to engage in conversations with humans and other agents, enabling them to learn and negotiate constraint costs more effectively. \\ \hline
\textbf{A3:} \raggedright Algorithm Simulating DCOP Agents & 
Rather than directly running a classical DCOP algorithm, A3 models the iterative coordination process as a sequential decision-making problem and uses an LFM-based policy to solve it. This allows A3 to simulate any coordination algorithm while also handling exceptional cases. \\ \hline
\end{tabular}
}
\caption{Spectrum of VL-DCOP Agents.}
\label{tab:SVD}
\end{table}

\begin{figure*}[t!]
    \centering
    \includegraphics[width=0.9\textwidth]{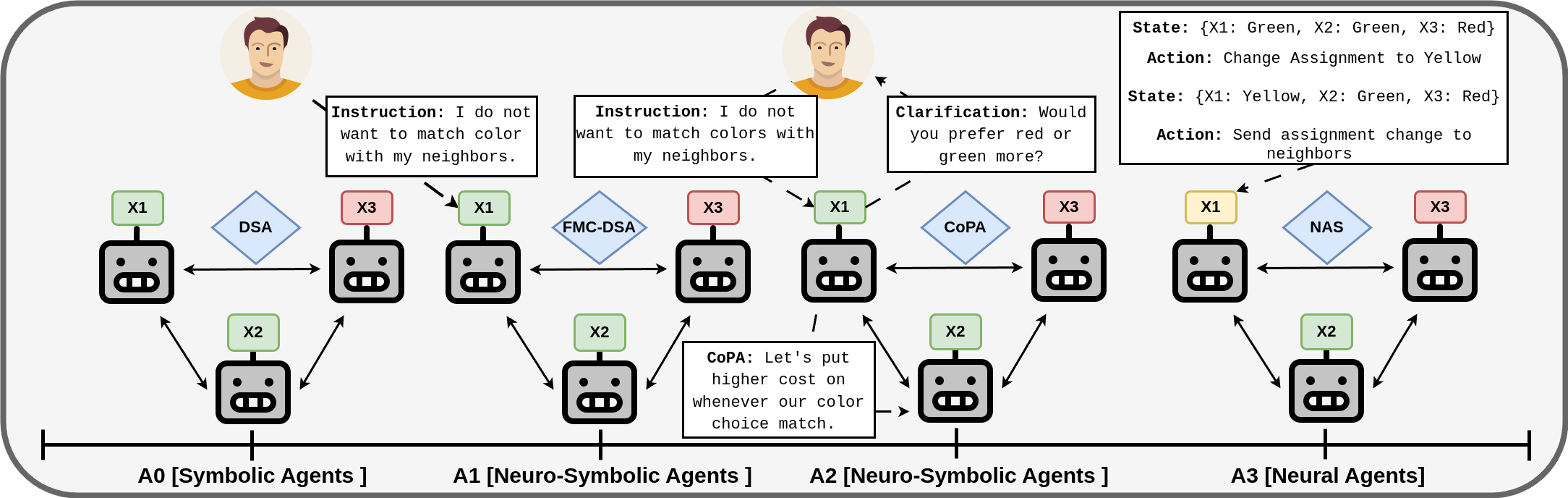}
    \caption{Spectrum of VL-DCOP Agents solving language-conditioned weighted graph coloring problems.}
    \label{fig:VLD}
\end{figure*}

To address these challenges, we introduce visual-linguistic instruction-based DCOPs, {VL-DCOPs}, an extended DCOP framework that models the interaction between humans and agents to specify constraints to solve a coordination task (Fig.~\ref{fig:VLD}). As a motivating example, consider a group of smartphone-based LFM-driven agents tasked with scheduling meetings for their owners. Initially, the agents use the instructions and contact lists provided to establish constraints with other agents. The LFM agents then coordinate by asking questions about schedules, reducing any uncertainty about the underlying cost structure. Since these agents rely on natural language, human responses can go beyond purely numerical feedback, incorporating explanations and other contextual details. Once constraints are established, the agents initiate a suitable optimization protocol until a mutually agreeable schedule is found. 

To implement the VL-DCOPs we design a spectrum of autonomous agent archetypes (Table~\ref{tab:SVD}) each utilizing a large multi-modal foundation model~\cite{Bhambri2023IncorporatingHF} to a varying degree. At one end of the spectrum, we have \textbf{A1} agents, a form of neuro-symbolic agent~\cite{sheth2023neurosymbolic} that uses these LFMs to parse the human instruction to make key algorithmic decisions such as action selection. However, the agent's workflow is still dictated by the classical algorithm. In this paper, we propose Foundation Model Centered DSA, \textbf{FMC-DSA}, a variant of the popular DSA~\cite{fitzpatrick2003distributed} algorithm for this archetype. 

Next, we present a more involved neuro-symbolic agent, \textbf{A2}, to address issues related to uncertainty about constraint cost due to natural ambiguity in linguistic communication and conflicts that may arise due to varying preferences among humans. To that end, we propose Cooperative Preference Aggregation (\textbf{CoPA}), an algorithm that uses LFMs to learn a good representation of the constraint function, and then uses a classical algorithm to solve the problem. Compared to A1, A2 provides a more interactive solution through conversation with humans and other agents.

Finally, we present \textbf{A3}, a general agent that in theory can simulate any algorithm depending on the instruction provided in the prompt. We model A3 as a sequential decision-making agent that uses an LFM-based (in-context) policy to solve the coordination problem. Unlike the previous two agents, for A3, we do not hard code any particular optimization procedure, making it a fully neural agent. The key motivation for this agent archetype is to provide flexibility to handle exceptional situations that may arise during optimization.     

To evaluate these agents, we propose three new benchmarks. The first two are based on classical distributed weighted graph coloring problems~\cite{Fioretto2016DistributedCO}. We design a linguistic (LDGC) and a visual (VDGC) variant of this problem to evaluate these agents. The third benchmark is based on more realistic meeting scheduling problems~\cite{Fioretto2016DistributedCO}. To build these agents we used pre-trained LLMs and VLMs such as LLAMA~\cite{dubey2024llama}, Qwen~\cite{qwen}, and GPT-4o\cite{hurst2024gpt}. Our analysis provides insight into the capability of current LFMs in distributed coordination and the advantages and disadvantages of each archetype. To our knowledge, our work is the first of its kind study in the DCOP literature and opens the door to many future research directions with interesting potential in explainability~\cite{edcop}, privacy~\cite{tassa2019privacy}, and exception handling such as message delay~\cite{rachmut2024latency,matsui2024investigation}. Further, with the rapid progress of GPU-based edge devices (such as Nvidia Jetson Orin Nano Super~\cite{nvidiaJetsonOrinNano}), and the ever-expanding capability of the frontier LFMs~\cite{llama3.2}, we expect a practical deployment of A3-like agents to be operational in the near future.

\section{Related Work}
\paragraph{DCOPs Framework}
As DCOPs must be designed manually, they face flexibility challenges in dynamic environments and scenarios where critical information is missing or difficult to obtain. Several extensions have been proposed to address this problem, including D-DCOP and U-DCOP~\cite{ddcop,Laut2011DistributedCO}. Dynamic-DCOPs (D-DCOPs) account for DCOPs that evolve over time, although these adaptations remain manually specified. U-DCOP offers a different perspective by employing probability distributions over constraint costs---manually learned from data---to enhance the framework's adaptability. In contrast, the proposed VL-DCOP can dynamically incorporate human instructions to construct new DCOP instances.

In parallel, a substantial body of literature on preference elicitation has investigated user queries as a means to reduce uncertainty in cost functions, thereby introducing additional adaptability~\cite{PE2,Tabakhi2017PreferenceEF,tsouros2024learning,exante,PEBC}. However, many such methods employ queries in a manually defined template and constrain the response options, often to a single numeric cost value. This may not be the most effective approach to communicating user preferences~\cite{tversky1982judgment}. Unlike these works, our research explores natural linguistic and visual mechanisms to express preferences. Moreover, the proposed CoPA algorithm further investigates agent-to-agent communication in natural language.

\paragraph{DCOPs Algorithm}
Over the last two decades, several algorithms have been proposed to solve DCOPs including, SyncBB~\cite{Hirayama1997DistributedPC}, DPOP~\cite{Petcu2005ASM}, DSA~\cite{fitzpatrick2003distributed}, MGM~\cite{Maheswaran2004Distributed}, ACO-DCOP~\cite{chen2018ant}, DPSA~\cite{mahmud2020learning}, Max-Sum~\cite{farinelli2008decentralised}, CoCoA~\cite{CoCoA}. In this paper, we focus on DSA, a widely recognized algorithm that captures the fundamental principles of DCOPs. While we acknowledge that the discussed agent archetypes can be applied to more advanced DCOP algorithms, we confine our attention to DSA in this work to accommodate space limitations.

\paragraph{LFM Centerd Decision Making}
Previous studies have explored agents that use large foundation models (LFMs)~\cite{cao2024survey} to process natural language instructions for sequential decision making. In our work, the A0 agent similarly makes single-step decisions from natural language inputs, but it is integrated with DSA for distributed optimization. The A3 agent can be considered a sequential decision-making agent; in our framework, it specifically decides on algorithmic steps to simulate distributed optimization. In parallel, other research has employed conversational agents~\cite{piriyakulkij2023active,Mahmud2024MAPLEAF} to infer user preferences, akin to the A2 agent in our approach using the proposed CoPA algorithm. However, these previous efforts typically focus on single-agent settings, whereas we apply CoPA to preference aggregation~\cite{prefag} in multi-agent scenarios. Furthermore, some recent work investigates a centralized Multi-Agent Debate (MAD) framework to learn a preference function~\cite{zou2024gradual}, whereas our CoPA method introduces a decentralized cooperative procedure to learn and consolidate preferences.

\paragraph{LFM Aided Constraint Programming}
Recent studies have utilized large foundation models (LFMs) for constraint programming~\cite{michailidis_et_al,tsouros2023holy,pan2023logic,Szeider2024MCPSolverIL,lawless2024want,regin2024combining,ye2024satlm}, addressing both modeling and solving tasks. However, these approaches predominantly involve centralized single-agent systems. While the A1 agent in our framework can be considered a multi-agent counterpart to such systems, to the best of our knowledge, existing works do not explore agent architectures akin to the A2 or A3 agents proposed in this study.
  
\paragraph{LFM Centered Agent Modeling}
Recent years have witnessed a growing interest in LFM-centered agent modeling across a variety of tasks~\cite{li2024review}. A notable example is the use of LFMs to simulate a complex distributed smart grid~\cite{amazonMAS}. While this approach resembles the A1 agent in our framework, it does not incorporate a systematic coordination algorithm such as FMC-DSA. Finally, previous studies, notably~\cite{schuurmans2023memory}, have demonstrated that memory-augmented LLMs are Turing complete. The A3 agent archetype in our work can be viewed as a practical implementation of this concept and, in theory, is capable of simulating any coordination algorithm given finite inputs and steps.

\section{Background}
Formally, a DCOP is defined by a tuple $ \langle \mathcal{A}, X, D, F,\delta \rangle $ \cite{modi2005adopt}. Here, $\mathcal{A}$ is a set of agents $\{a_1, a_2, ..., a_n\}$, and $X$ is a set of discrete variables that are controlled by this set of agents. $D$ is a set of discrete domains and $D_i \in D$ is a set that contains values that may be assigned to its associated variable $x_i$. $F$ is a set of constraints where $f_i \in F$ is a function of a subset of variables $x^i \subseteq X$. Thus, the function $f_i : \times_{x_j \in x^i} D_j \to \!R $ denotes the cost for each possible assignment of the variables in $x^i$. $\delta: X \rightarrow A$ is a variable-to-agent mapping function that assigns the control of each variable $x_i \in X$ to an agent of $A$. Each variable is controlled by a single agent. However, each agent can hold several variables. Within the framework, the objective of a DCOP algorithm is to produce $X^*$, a complete assignment that minimizes the aggregated cost of the constraints, defined below:
\begin{equation}
    X^* = \argmin_X \sum_{i=1}^{l} f_i(x^i)
    \label{eqobj}
\end{equation}

The Distributed Stochastic Algorithm (DSA) is a widely used approach for solving Distributed Constraint Optimization Problems (DCOPs). Each agent begins with a random assignment and engages in an iterative optimization process. During each iteration, the agent communicates its assignment to neighbors, updates its local context based on their messages, and computes the locally optimal assignment. This assignment is adopted probabilistically, following a stochasticity parameter \(\epsilon\). This \(\epsilon\)-greedy strategy helps DSA agents to escape local minima. The process repeats until convergence, providing an efficient framework for distributed optimization.

\begin{algorithm}[t]
\caption{Distributed Stochastic Algorithm (DSA)}
\small
\begin{algorithmic}[1]
\REQUIRE Stocasticity parameter, $\epsilon$
\STATE $\mathcal{C}_{actions} \leftarrow \text{Initialize local assignment context randomly}$
\WHILE{not converged}
    \STATE Send the current assignment to all neighbors
    \STATE Receive messages from neighbors and update $\mathcal{C}_{actions}$ 
    \STATE $d_{\text{max}} \leftarrow$ Least local cost action given   $\mathcal{C}_{actions}$
    \STATE Change current assignment to $d_{\text{max}}$ with probability $1 {-} \epsilon$ or to a random assignment with probability $\epsilon$
\ENDWHILE
\end{algorithmic}
\end{algorithm}

\section{Details of VL-DCOPs}

We formally define VL-DCOPs as a tuple $(\mathcal{A}^C,\mathcal{A}^I, \Gamma, E, \mathcal{T}_t)$ where:

\begin{itemize}

    \item $\mathcal{A}^C$ is a set of agents, analogous to agents in DCOPs, that will take part in coordination tasks.
    
    \item $\mathcal{A}^I$ is a set of instructing agents that will provide task descriptions to agents in $\mathcal{A}^C$. For example, it can be a set of human providing instruction to their virtual assistants ($\mathcal{A}^C$) or a set of LFMs that are providing instruction to another set of coordinating LFMs ($\mathcal{A}^C$). For some problems, the same LMF can play different roles.
    
    \item $\Gamma$ is a mapping function that maps agent in $\mathcal{A}^C$ to agents in $\mathcal{A}^I$ to define flow of instruction. In practice, this may be implicitly defined; for example, by default, the owner of a mobile device has control over the virtual agent on the device.  
    
    \item $E$ defines the network structure\footnote{Can be thought of as a set of edges.} of the agents. This is not directly analogous to $F$ in DCOP as constraints are not established a priori. This only defines which agent is allowed to communicate with other agents. 
    
    \item $\mathcal{T}_t$ is the task at the $t$-{th} moment, defined by the tuple $(\mathcal{I},X, D, F)$. Here, $\mathcal{I}$ is a set of local visual-linguistic instructions that provide a local description of the coordination task. For a simple example, in a linguistic graph coloring problem (see Section\,\ref{sect:experiments} for details) a human might provide available colors, preference order of each color, and neighbors with which they don't want to match color. $X, D, F$ are analogous to DCOPs.  
    
\end{itemize}
Note that this is the general description of VL-DCOPs and a practical implementation of the VL-DCOPs instance might not need all the components. The objective at time $t$-th time still remains the same as DCOP. However, it is important to note that the VL-DCOP agents may only observe $\mathcal{I}$ directly and then infer (explicitly or implicitly) $X, D,$ and $F$ from $\mathcal{I}$. So in practice, they would often optimize an approximation of $\mathcal{T}_t$.       
\section{Agents for VL-DCOPs}
In this section, we provide details of the three agent archetypes that we briefly introduced to implement VL-DCOPs.

\subsection{A1: Instruction Following DCOP Agent}
We begin by describing an agent designed to solve a simple instantiation of VL-DCOPs. Specifically, for each element $E$, there is a corresponding $F$, and the variables $X$ and the corresponding domains $D$ are fixed. $A1$ is a basic LFM-centered agent that processes visual-linguistic instruction $\mathcal{I}_i$ as input and translates it into a decision while coordinating with other agents. To achieve this, we propose FMC-DSA, a straightforward extension of DSA tailored for this purpose.

\begin{algorithm}[t]
\caption{Foundation Model Centered DSA (FMC-DSA)}
\small
\begin{algorithmic}[1]
\REQUIRE Stochasticity parameter, $\epsilon$, Foundation Model, $\mathcal{M}$, Task Instruction, $\mathcal{I}_i$  
\STATE $\mathcal{C} \leftarrow \text{Initialize local assignment context randomly}$
\FOR{each $n_j$ in neighbors}
    \STATE $m_{i,\bar{j}} \leftarrow \mathcal{M}(\mathcal{I}_i, n_j, \text{GENERATE CONSTRAINT})$
    \STATE Send $m_{i,\bar{j}}$ to $n_j$
\ENDFOR
\STATE Receive preference messages $\{m_{\bar{i},j}\}$ from all neighbors
\STATE $\mathcal{O} \leftarrow \{\mathcal{I}_i\} \cup \{m_{\bar{i},j}\}$
\WHILE{not converged}
    \STATE Send the current action to all neighbors
    \STATE Receive messages from neighbors and update $\mathcal{C}$ 
    \STATE $d_{\text{max}} \leftarrow \mathcal{M}(\mathcal{C}, \mathcal{O}, \text{GET MAX ACTION})$ 
    \STATE Change the current assignment to $d_{\text{max}}$ with probability $1 - \epsilon$, or to a random assignment with probability $\epsilon$
\ENDWHILE
\end{algorithmic}
\end{algorithm}

The FMC-DSA agent begins by randomly initializing the local variable assignment $\mathcal{C}$. It then generates natural language constraint descriptions with the help of the LFM $\mathcal{M}$ for each of its neighbors and sends these as constraint initialization messages, denoted $m_{i,\bar{j}}$. Similarly, the agent receives messages from all neighbors and stores them, along with their original instruction, in the set $\mathcal{O}$ (Algorithm 2, lines $2-7$). 

The agent iteratively performs two steps until the termination criterion (e.g., the maximum number of iterations) is met:

\begin{enumerate}
    \item It sends and receives the current assignment from neighbors, updating the local context $\mathcal{C}$ accordingly (Algorithm 2, lines $9-10$).
    \item The agent prompts the LFM $\mathcal{M}$ to generate the optimal assignment based on the current assignment context $\mathcal{C}$ and the set $\mathcal{O}$. The agent then updates its assignment to the best option with probability $1 - \epsilon$ or randomly with probability $\epsilon$ (Algorithm 2, lines $11-12$).
\end{enumerate}
This can be viewed as a local $\epsilon$-greedy strategy that facilitates coordination among the agents using LFMs. However, it is important to note that FMC-DSA introduces preference asymmetry, which can impact the coordination process. Specifically, although agents locally exchange information about constraints, they do not reach a consensus on the cost of each assignment. As a result, two coordinating agents may interpret the cost of the same assignment differently. We will address this issue in the next sub-section.   
\subsection{A2: Conversational DCOP Agent}
Consider the LGCP problem scenario where agent X receives the instruction to avoid the color of agent Y, and the general color preference is $A \succ B \succ C$. Agent Y is instructed to avoid the color of agent X, and their preference is $B \succ A \succ C$. Therefore, according to agent X, the cost order is $(C,C) > (A,A) > (B,B)$, while according to agent Y, the cost order is $(C,C) > (B,B) > (A,A)$. In FMC-DSA, while the agent recognizes this difference, there is no coordination between the agents to reach a consensus about the assignment preference order. To address this, we now introduce an instance of A2, where agents go through a discussion process to achieve a consensus, called Cooperative Preference Aggregation (CoPA).

\begin{algorithm}[t]
\caption{Cooperative Preference Aggregation (CoPA)}
\small
\begin{algorithmic}[1]
\REQUIRE Number of rounds $R$, Foundation Model, $\mathcal{M}$, Task Instruction, $\mathcal{O}$ 
\STATE $F^0_{i,\bar{j}} \leftarrow \mathcal{M}(\mathcal{I}_i, \text{GENERATE CONSTRAINT})$
\STATE $\mathcal{H} \leftarrow \{F^0_{i,\bar{j}}\}$
\FOR{$r \in \{1, 2, \dots , R\}$}
    \STATE Send $F^{r-1}_{i,\bar{j}}$ to neighbor $j$
    \STATE $F^{r-1}_{\bar{i},j} \leftarrow$ Receive from neighbor $j$
    \STATE $F^r_{i,\bar{j}} \leftarrow \mathcal{M}(\mathcal{I}_i, \mathcal{H}, F^{r-1}_{\bar{i},j}, \text{GENERATE CONSTRAINT})$
    \STATE $\mathcal{H} \leftarrow \mathcal{H} \cup F^{r-1}_{\bar{i},j} \cup \{F^r_{i,\bar{j}}\}$
\ENDFOR
\STATE $F^R_{i,j} \leftarrow \mathcal{M}(F^R_{i,\bar{j}}, F^R_{\bar{i},j}, \text{RESOLVE})$
\end{algorithmic}
\end{algorithm}

CoPA is an iterative discussion process used to generate a cost table. The procedure starts after constructing $\mathcal{O}$, similar to FMC-DSA (which is omitted here). Initially, both agents construct constraints. They then iteratively propose the cost table to their neighbor. Based on the two cost tables, they redesign the cost structure and propose the updated structure for $K$ rounds. In the end, if a consensus is reached, they can use the agreed-upon cost function; otherwise, the conflict is resolved based on a heuristic, such as taking the average or maximum.

For practical implementation, in order to keep the cost consistent across different agents, one might prompt the agents to keep it within the range $[Cost_{\text{min}}, Cost_{\text{max}}]$. Additionally, ambiguity may arise from the natural language instructions themselves, and additional queries may be required from the instruction-giving agents (\(\mathcal{A}^I\)) (during or before CoPA). However, prior work~\cite{Mahmud2024MAPLEAF} exists that falls under the A2 archetype, and can be directly applied. Due to space constraints, we do not explore these alternative A2 DCOP agents.

\subsection{A3: Algorithm Simulating DCOP Agent}

The A3 agent models the iterative coordination process as a sequential decision-making problem, formulated as a Markov Decision Process (MDP). This framework allows the agent to simulate classical coordination algorithms while dynamically adapting them online to handle exceptional cases.

\paragraph{MDP Definition:}
An MDP for simulating algorithms can be defined as $\langle \mathcal{S}, \mathcal{A}, \mathcal{T}, \mathcal{R}\rangle$:
\begin{itemize}
    \item $\mathcal{S}$: The state space, represented by the history of algorithm execution log.
    \item $\mathcal{A}$: The action space, encompassing decisions like action selection, message passing, and other function calls.
    \item $\mathcal{T}$: The transition function, which updates the algorithmic log based on actions and interactions with neighbors (assumed model-free).
    \item $\mathcal{R}$: The reward function, representing local or global costs (if anytime mechanisms exist).
\end{itemize}

Instead of explicitly training a policy from scratch, we leverage Large Foundation Models (LFMs) as generalized policies that rely on in-context learning. The LFM interprets the algorithmic log and determines the next action. Prompts for the LFM may include descriptions of various DCOP algorithms, but our observations suggest that many LFMs, due to their pre-training data, already possess knowledge of common DCOP methods. Algorithm~\ref{alg:in_context_simulation} provides a general outline of this approach.

In this setup, $ENV$ represents the interaction interface through which the agent communicates with other agents and its own systems. The interface continuously monitors the environment and observes task initialization $\mathcal{T}_t$, incoming messages, and internal system states.

\begin{algorithm}[t]
\caption{Neural Algorithm Simulation}
\label{alg:in_context_simulation}
\small
\begin{algorithmic}[1]
\REQUIRE Initial instruction $\mathcal{P}$, Foundation Model $\mathcal{M}$, Interaction Interface $ENV$
\STATE Initialize algorithmic log $\mathcal{L}$ with $\mathcal{P}$
\FOR{$t \in 1, 2, \dots$}
    \STATE $a_t \gets \mathcal{M}(\mathcal{L}, \text{GET ACTION})$
    \STATE $o_t \gets ENV(a_t)$
    \STATE Update algorithmic log $\mathcal{L}$ with action $a_t$ and observation $o_t$
\ENDFOR
\end{algorithmic}
\end{algorithm}

\paragraph{Simulating FMC-DSA with A3}
We now provide an example of the A3 agent simulating FMC-DSA by iteratively updating its algorithmic log:
\begin{itemize}
    \item \textbf{Initialization:} The initial task instruction is observed through $E$ as $o_t$ at time $t$. Before this, the agent may be engaged in other tasks. The agent starts with an initialization action that generates random assignments and constraints, which are then added to the log $\mathcal{L}$.
    \item \textbf{Message~Passing:} The agent performs a series of message-passing actions, receiving preferences (e.g., $m_{\bar{i},j}$) from neighbors, and updates the context and log accordingly.
    \item \textbf{Assignment~Updates:} The agent alternates between assignment changes and message-passing actions, iteratively refining its decisions.
    \item \textbf{Interruption:} If an error or interruption occurs during optimization, the A3 agents will adaptively respond based on their policy.
\end{itemize}

A key advantage of this approach is its ability to adapt to exceptional scenarios in real-time. Some examples include: If a message fails to arrive, the agent can continue optimization with partial information. The agent can also make meta-algorithmic decisions such as when to commit to an action and hyper-parameter selection, adjust constraints mid-optimization, or incorporate dynamic changes in the task environment. This flexibility makes A3 robust in environments with uncertainties or dynamic requirements, while still adhering to the principles of classical DCOP coordination algorithms.

\section{Analysis}
This section will analyze the advantages and disadvantages of different agent archetypes. 

\paragraph{Query Complexity:} The primary computational cost in VL-DCOPs algorithms arises from the number of LLM queries required. \textbf{A1} agents make one LLM query per iteration, resulting in $O(n)$ queries where $n$ is the number of iterations. In contrast, \textbf{A2} agents make $O(k)$ queries per neighbor, where $k$ is the number of CoPA rounds, requiring a total of $O(k\cdot Deg)$ queries, where $Deg$ represents the average degree of the graph. Since $k \ll n$ is in practice, A2 agents generally perform fewer queries in sparse graphs. On the other hand, \textbf{A3} agents require $O(c\cdot n)$ LLM queries, where $c$ represents the average number of algorithmic decisions made per iteration. In practice, A3 agents are the most computationally expensive VL-DCOP agents.

\begin{table*}[ht]
    \footnotesize
    \centering
    \renewcommand{\arraystretch}{1.0}
    \setlength{\tabcolsep}{3pt}
    \begin{tabular}{@{}l|l|cccc|cccc|cccc@{}}
        \toprule
        \multirow{3}{*}{\textbf{Model Name}} & \multirow{3}{*}{\textbf{Agent Type}} & \multicolumn{12}{c}{\textbf{Benchmark Performance}} \\
        \cmidrule{3-14}
        & & \multicolumn{4}{c|}{\textbf{LDGC}} & \multicolumn{4}{c|}{\textbf{VLDGC}} & \multicolumn{4}{c}{\textbf{LDMS}} \\
        \cmidrule{3-14}
        & & \textbf{Cost} & \textbf{Anytime} & \textbf{Sat.} & \textbf{Anytime} & \textbf{Cost} & \textbf{Anytime} & \textbf{Sat.} & \textbf{Anytime} & \textbf{Cost} & \textbf{Anytime} & \textbf{Sat.} & \textbf{Anytime} \\
        \midrule
        \multirow{3}{*}{Qwen 2.5 3B} & FMC-DSA & $117.8$ & $82.7$ & $63.2\%$ &$76.6\%$ & $152.2$ & $109.3$ & $50.9\%$ &$60.8\%$ & $69.1$ & $65.5$ & $19.2\%$ & $28.7\%$ \\
        & CoPA+DSA &  $160.2$ & $160.5$ & $50.2\%$ &$50.5\%$ & $163.2$ & $155.1$ & $49.6\%$ &$69.1\%$ & $39.7$ & $50.8$  & $26.1\%$ & $28.2\%$ \\
        & NAS & $159.2$ & $122.5$ & $50.5\%$ &$60.5\%$ & $158.2$ & $123.5$ & $55.5\%$ &$54.7\%$ & $65.6$ & $48.2$ & $19.8\%$ & $10.3\%$ \\
        \midrule
        \multirow{3}{*}{Llama 3.3 70B} & FMC-DSA & $95.3$ & $71.5$ & $65.1\%$ &$79.6\%$ & $100.5$ & $72.6$ & $64.6\%$ &$77.4\%$ & $54.8$ & $39.6$  & $23.0\%$ & $38.2\%$ \\
        & CoPA+DSA & $103.3$ & $82.7$ & $74.4\%$ &$81.1\%$ & $102.2$ & $\mathbf{84.5}$ & $74.6\%$ &$81.4\%$ & $48.0$ & $\mathbf{36.2}$  & $30.8\%$ & $38.2\%$ \\
        & NAS & $159.9$ & $122.1$ & $50.2\%$ &$50.5\%$ &  $159.8$ & $121.1$ & $50.8\%$ &$60.2\%$ & $65.1$ & $48.5$  & $12.1\%$ & $19.6\%$ \\
        \midrule
        \multirow{3}{*}{GPT-4o} & FMC-DSA & $100.5$ & $66.1$ & $70.8\%$ &$83.5\%$ & $103.5$ & $66.4$ & $71.1\%$ &$82.6\%$ & $41.4$ & $31.5$  & $38.0\%$ & $50.4\%$ \\
        & CoPA+DSA & $97.4$ & $\mathbf{74.2}$ & $73.9\%$ &$81.9\%$ & $106.1$ & $93.7$ & $75.1\%$ &$81.4\%$ & $47.4$ & $75.1$  & $81.3\%$ & $41.1\%$\\
        & NAS & $103.7$ & $73.5$ & $62.8\%$ &$71.1\%$ & $114.6$ & $72.1$ & $61.8\%$ &$73.6\%$ & $45.1$ & $33.7$  & $39.2\%$ & $41.6\%$ \\
        \midrule
        \multirow{3}{*}{GPT-4o-mini} & FMC-DSA & $\mathbf{88.1}$ & $65.4$ & $67.9\%$ &$76.0\%$ & $\mathbf{90.2}$ & $58.2$ & $68.1\%$ &$82.6\%$ & $\mathbf{40.1}$ & $30.3$  & $39.6\%$ & $53.2\%$ \\
        & CoPA+DSA & $111.5$ & $99.8$ & $65.2\%$ &$66.1\%$ & $128.2$ & $134.3$ & $61.7\%$ &$65.6\%$ & $47.7$ & $36.3$  & $32.6\%$ & $41.3\%$ \\
        & NAS & $\mathbf{89.5}$ & $67.3$ & $68.6\%$ &$74.2\%$ & $\mathbf{91.5}$ & $62.2$ & $67.5\%$ &$82.1\%$ & $\mathbf{41.1}$ & $33.5$  & $40.3\%$ & $52.3\%$ \\
        \midrule
        \midrule
        \multirow{2}{*}{\textcolor{black}{\textbf{Oracle}}} & DSA & $74.9$ & $55.6$ & $74.7\%$ & $84.3\%$ & $68.4$ & $49.5$ & $74.5\%$ & $84.0\%$ & $48.1$ & $36.7$ & $39.4\%$ & $52.8\%$ \\
        & Random & $161.1$ & $101.3$ & $50.2\%$ & $71.5\%$ & $159.5$ & $98.5$ & $51.7\%$ & $73.3\%$ & $67.1$ & $49.7$ & $7.8\%$ & $22.4\%$ \\
        \bottomrule
    \end{tabular}
    \caption{Performance of different models and agents across benchmarks with baseline comparisons.}
    \label{tab:model_performance}
\end{table*}

\label{sect:experiments}
\paragraph{Benchmarks:} 
We present three different benchmarks to evaluate VL-DCOP. The first is \textbf{LDGC}, a language-based weighted graph coloring problem. Here, each agent is instructed in natural language about which agents they should match colors with and which agents they should avoid matching colors with. In addition, instructing agents (\(\mathcal{A}^I\)) describe their color preference order using natural language. The second benchmark is \textbf{VLDGC}, which incorporates both visual and language inputs. Specifically, the color preferences are presented using various types of plots, such as bar charts, line charts, and histograms, while the color-matching constraints are described in natural language. Finally, \textbf{LDMS} addresses the classical meeting scheduling problem. However, in this benchmark, the time slot preferences for each meeting are described using natural language. A key difference from the previous two benchmarks is that, in LDMS, the agent must make decisions involving multiple variables, as opposed to the single-variable focus of the others.

\paragraph{Experimental Setup:} 
We test three open-sourced pretrained foundation models in our experiments: \textbf{LLAMA 3.3 70B 4-bit Instruct}~\cite{dubey2024llama}, representing larger foundation models; \textbf{Qwen 2.5 3B Instruct}~\cite{qwen}, representing smaller foundation models that can run on mobile devices; and \textbf{ModernBERT Large 0.4B}~\cite{modernbert}, suitable for edge devices with NPUs or CPUs. Additionally, we evaluate \textbf{GPT-4o} and \textbf{GPT-4o-mini}, providing insight into the capabilities of state-of-the-art (SOTA) foundation models. To process visual data, we use the \textbf{LLAMA 3.2 11B} model paired with various language models. For the A3 agent, we created a Gymnasium-style scaffolding, where these LFM-based agents control virtual DSA agents to simulate DSA algorithms. The experiments were conducted on a desktop with \textbf{128 GB RAM}, an \textbf{Intel Xeon w3-2423} CPU, and an \textbf{Nvidia A6000 ADA 48GB GPU}. All experiments are reproducible with a computational budget of approximately \textbf{\$200 USD}.

\subsection{Agent Archetype Analysis}
In this subsection, we analyze the agent archetypes where \textbf{A1} is represented by FMC-DSA, \textbf{A2} is represented by CoPA+DSA ($k=2$), and \textbf{A3} is represented by NSA. For each benchmark, we generated 10 different problem instances, each with 10 agents and 23 edges, with domain size 4. Each agent was simulated for 50 iterations. Additionally, we present results for DSA and random baselines simulated using a handcrafted cost function based on the instructions for 100 iterations to provide a comparison with ground-truth costs. We set $\epsilon = 0.1$ for all cases except for $A3$ which was set to $0.03$ as simulation inaccuracy introduced a second source of stochasticity.  The \textit{cost} represents the total mean cost for the last 50\% of the iterations. The \textit{anytime cost} is the best solution found across all iterations. \textit{Sat} is the number of constraints satisfied without considering the preference cost. The costs of the assignments are calculated using a handcrafted cost table.

We observe the trend that A1 agents achieve better anytime costs than A2 agents. However, it is important to note that A1 agent networks, in practice, cannot use the anytime mechanism~\cite{zivan2008anytime} as they cannot calculate the global cost. However, in theory, A1 agents can find better solutions than A2 agents. This indicates for LFMs the task of inferring cost tables is more difficult than the assignment selection task. 

A2 agents generally perform slightly worse than A1 agents. Importantly, for A2 agents, it is possible to calculate the anytime cost. In such cases, A2 agents outperform A1's average cost across all tasks. This highlights the advantage of A2's ability to compute and utilize anytime costs. 

Finally, A3 agents perform similarly or slightly worse than A1 agents across all tasks, as they simulate A1 agents. The slight performance degradation arises from imperfect simulations. Most models struggled with running the simulations, with GPT-4o-mini being the only model capable of near-perfect simulations. We observed that the first 10 iterations were perfectly simulated, but the simulation quality later degraded, achieving only about 65\% correct algorithmic decisions. This suggests that A3 agents might be better suited for simulating non-iterative algorithms like CoCoA\cite{CoCoA}. These results also demonstrate the potential to design adaptive LFM-centered algorithms.

\begin{table}[t]
    \footnotesize
    \centering
    \renewcommand{\arraystretch}{1.0}
    \setlength{\tabcolsep}{2.5pt}
    \begin{tabular}{@{}l|l|cccc@{}}
        \toprule
        \multirow{2}{*}{\textbf{Model Name}} & \multirow{2}{*}{\textbf{Agent Type}} & \multicolumn{4}{c}{\textbf{LDGC Benchmark Performance}} \\
        \cmidrule{3-6}
        & & \textbf{Cost} & \textbf{Anytime} & \textbf{Sat.} & \textbf{Anytime} \\
        \midrule
        \multirow{3}{*}{GPT-4o-mini} & FMC-DSA & $512.6$ & $445.2$ & $61.2\%$ & $68.7\%$ \\
        & CoPA+DSA & $624.4$ & $576.1$ & $62.3\%$ & $66.6\%$ \\
        & NAS & $521.1$ & $465.2$ & $60.6\%$ & $68.5\%$ \\
        \midrule
        \multirow{3}{*}{ModernBART 0.4B} & FMC-DSA &$420.2$ & $375.1$ & $73.1\%$ & $77.6\%$ \\
        & CoPA+DSA & $421.1$ & $370.9$ & $73.6\%$ & $75.9\%$ \\
        & NAS &$419.6$ & $372.8$ & $73.3\%$ & $77.8\%$ \\
        \midrule
        \midrule
        \multirow{2}{*}{\textcolor{black}{\textbf{Oracle}}} & DSA & $422.5$ & $370.2$ & $73.1\%$ & $77.8\%$ \\
        & Random & $849.1$ & $747.14$ & $49.3\%$ & $56.6\%$ \\
        \bottomrule
    \end{tabular}
    \caption{Performance on the large random network.}
    \label{tab:model_performance}
\end{table}

GPT-4o-mini outperformed other models, including larger GPT-4o, for A1 and A3 agents. However, for A2 agents, both GPT models underperformed compared to LLAMA models in VLDGC and LDMS. LLAMA 3.3 70B achieved the best performance for A2, and as A2's anytime cost is representative of real-world performance, it outperformed all other models. None of the open-sourced models resulted in a capable A3 agent. Additionally, the Qwen model performed similarly to the random baseline, except for the LDGC A1 task. The main takeaway is that these models, even without any task-specific training, can solve VL-DCOP tasks. However, there is room for significant improvements.

\subsection{Scaling Up the Benchmark}

In this subsection, we scale up the benchmark and evaluate performance using 10 instances of the LDGC problem, each consisting of 50 agents, 120 edges, and a domain size of 4. We consider two types of network architectures: \textbf{random} and \textbf{scale-free}~\cite{barabasi1999emergence}.

We use the \textbf{GPT-4o-mini} model, which demonstrates excellent performance and cost-effectiveness. However, the computational requirements and cost of other models make them prohibitively expensive to utilize at this scale. To address this, we consider the \textbf{ModernBART} Large model, a compact and efficient alternative model capable of running on CPUs and performing significantly faster than OpenAI API calls when deployed on GPUs. This efficiency makes it suitable for practical applications and deployment on edge devices, such as Raspberry Pi. 

The main limitation of ModernBART and similar small models is their lack of general intelligence and inability to handle general decision-making tasks. As a result, these models cannot be directly used to implement the proposed agent without modifications. To overcome this limitation, we perform task-specific training on ModernBART. Specifically, we use ground truth decisions generated by the DSA algorithm along with prompts from the GPT-4o-mini model as the training dataset. For this purpose, we collect 8,000 decision pairs for each type of agent. Overall, this training process requires less than 2 hours to complete.

\textbf{Performance:} ModernBERT achieves perfect task performance, surpassing even the GPT models in accuracy when evaluated against DSA. The key advantages are as follows: \textbf{VL-DCOP} can be deployed on edge devices, enabling real-time operations in distributed networks. The model demonstrates high scalability, with simulations of 50 A1 agents over 50 iterations completed in less than 3 minutes on a single GPU system. We also found that A2 agents were more query-efficient and quicker to simulate on these sparse graphs. However, there are notable disadvantages: ModernBERT is not a general-purpose \textbf{VL-DCOP} agent and is limited to solving specific tasks for which it was trained. The trained models perform well only when input prompts are similar in length and structure to the original training dataset. Performance degrades significantly when there is considerable dissimilarity. Overall, the findings indicate that \textbf{VL-DCOP} can be highly scalable and can be implemented on the current hardware, provided specific optimizations are implemented.

\begin{table}[t]
    \footnotesize
    \centering
    \renewcommand{\arraystretch}{1.0}
    \setlength{\tabcolsep}{2.5pt}
    \begin{tabular}{@{}l|l|cccc@{}}
        \toprule
        \multirow{2}{*}{\textbf{Model Name}} & \multirow{2}{*}{\textbf{Agent Type}} & \multicolumn{4}{c}{\textbf{LDGC Benchmark Performance}} \\
        \cmidrule{3-6}
        & & \textbf{Cost} & \textbf{Anytime} & \textbf{Sat.} & \textbf{Anytime} \\
        \midrule
        \multirow{3}{*}{GPT-4o-mini} & FMC-DSA & $515.3$ & $434.2$ & $62.1\%$ & $69.6\%$ \\
        & CoPA+DSA & $612.1$ & $512.9$ & $64.3\%$ & $67.8\%$ \\
        & NAS & $514.1$ & $438.5$ & $63.3\%$ & $69.0\%$ \\
        \midrule
        \multirow{3}{*}{ModernBART 0.4B} & FMC-DSA & $405.5$ & $352.1$ & $74.3\%$ & $79.5\%$ \\
        & CoPA+DSA & $401.5$ & $356.1$ & $74.2\%$ & $79.2\%$ \\
        & NAS & $409.5$ & $352.1$ & $74.1\%$ & $78.9\%$ \\
          \midrule
        \midrule
        \multirow{2}{*}{\textcolor{black}{\textbf{Oracle}}} & DSA & $403.5$ & $350.1$ & $74.6\%$ & $79.5\%$ \\
        & Random & $831.4$ & $714.3$ & $50.2\%$ & $57.0\%$ \\
        \bottomrule
    \end{tabular}
    \caption{Performance on the large scale-free network.}
    \label{tab:model_performance}
\end{table}

\section{Conclusion and Future Directions}

We introduce VL-DCOPs, a novel framework that integrates visual and linguistic instructions with DCOPs, making the system more dynamic and natural to use. We propose several LFM-centered agent archetypes to solve VL-DCOP tasks. Our experiments demonstrate that LFMs are capable of addressing these problems off-the-shelf and can be further fine-tuned to achieve performance comparable to symbolic systems. Our work opens up several promising avenues for future research:

\paragraph{Adaptive Algorithms:} By demonstrating the feasibility of using general algorithm-simulating agents in VL-DCOPs, this work paves the way for adaptive algorithms that can handle exceptions such as network delays and other types of interruptions. Investigating these adaptive capabilities is an important direction for future research.
    
\paragraph{Explainability and Interpretability:} Incorporating visual and linguistic understanding can make agents naturally more interpretable. Further work on improving and evaluating the explainability and interpretability of VL-DCOP agents would be highly valuable.

\paragraph{Privacy and Security Concerns:} There are several privacy and security issues inherent to VL-DCOP networks. For instance, malicious agents with general intelligence could influence other agents' decisions to favor their preferences, extract sensitive information about the network structure, and so on. Addressing these issues and exploring mechanisms to prevent such gaming scenarios is a crucial direction for future work.

Overall, our work highlights the potential of VL-DCOPs and the need for continued research into adaptive algorithms, interpretability, and privacy to fully realize their capabilities in dynamic, real-world scenarios.

\newpage

\bibliographystyle{named}
\bibliography{ijcai25}

\end{document}